\pdfoutput=1

\documentclass[11pt]{article}

\usepackage{acl}
\usepackage{amsmath}

\usepackage{times}
\usepackage{multirow}
\usepackage{xcolor}
\usepackage{tabularx,paralist}
\usepackage{enumitem}
\usepackage{array,booktabs,ragged2e}
\usepackage{algorithm}
\usepackage{algpseudocode}
\newcolumntype{R}[1]{>{\RaggedLeft\arraybackslash}p{#1}}
\usepackage{latexsym}
\usepackage{tabularx}
\usepackage{colortbl}

\usepackage[utf8]{inputenc}
\usepackage{graphicx}
\usepackage{microtype}

\newcommand*{\affmark}[1][*]{\textsuperscript{#1}}
\renewcommand\footnotemark{}
\title{Did You Read the Instructions? \\ Rethinking the Effectiveness of Task Definitions in Instruction Learning}
\author{Fan Yin\affmark[$*\S$]\thanks{$^*$Work done when Fan Yin was an intern at Salesforce.},~Jesse Vig\affmark[$\diamondsuit$]\affmark[$\dagger$],~Philippe Laban\affmark[$\diamondsuit$]\affmark[$\dagger$]\thanks{$^\diamondsuit$Jesse and Philippe contributed equally; order is random.}, \\  \bf Shafiq Joty\affmark[$\dagger$], Caiming Xiong\affmark[$\dagger$], Chien-Sheng Jason Wu\affmark[$\dagger$]\\
{\affmark[$\S$]UCLA}\quad {\affmark[$\dagger$]Salesforce AI Research}\\
\texttt{fanyin20@cs.ucla.edu} \\ \texttt{\{jvig, plaban, sjoty, wu.jason, cxiong\}@salesforce.com} 
}

\begin{document}
\maketitle

\begin{abstract}

Large language models (LLMs) have shown impressive performance in following natural language instructions to solve unseen tasks. However, it remains unclear whether models truly understand task definitions and whether the human-written definitions are optimal. In this paper, we systematically study the role of task definitions in instruction learning. We first conduct an ablation analysis informed by human annotations to understand which parts of a task definition are most important, and find that model performance only drops substantially when removing contents describing the task output, in particular label information. Next, we propose an automatic algorithm to compress task definitions to a minimal supporting set of tokens, and find that 60\% of tokens can be removed while maintaining or even improving model performance. Based on these results, we propose two strategies to help models better leverage task instructions: (1) providing only key information for tasks in a common structured format, and (2) adding a meta-tuning stage to help the model better understand the definitions. With these two strategies, we achieve a 4.2 Rouge-L improvement over 119 unseen test tasks.

\end{abstract}

\section{Introduction}

\begin{table*}[t]
\renewcommand{\arraystretch}{0.85}
\begin{center}
\small
\begin{tabular}{p{15.15cm}}
\toprule
\textbf{RQ1:} Which parts of task definitions are important when performing zero-shot instruction learning? \cr
\midrule
- For classification tasks, label-related information is crucial, as it helps the model identify the output space and identify each label's meaning when generalizing. \cr
- Additional details or constraints besides primary mentions of input and output, in general, do not improve model performance. As model size increases, additional details become important.\cr
- Task definitions can be extensively compressed with no performance degradation, particularly for generation tasks. \cr

\midrule
\textbf{RQ2:} Is natural language the most efficient format to communicate task instructions to models? \cr
\midrule 
- Framing instructions as a structured input/action/output triplet is potentially a more efficient and effective way of creating task definitions.\cr
- In fact, using only basic metadata and the label space (without label definitions) in a structured format, we achieve similar, or even better performance as with full definitions.\cr

\midrule
\textbf{RQ3:} How can we improve models' understanding of definitions as well as model performance? \cr
\midrule
 - Adding a meta-tuning stage for adapting models to the writing styles of definitions improves the performance. \cr
\bottomrule\hline
\end{tabular}
\vspace{-6pt}
\end{center}
\caption{Summary of research questions and key findings of the paper.}
\label{table:summaryoffindings}
\end{table*}

Large language models or LLMs~\citep{devlin2019bert, 2020t5, Brown2020LanguageMA} demonstrate the ability to perform zero-shot cross-task generalization through learning from instructions of tasks~\citep{Sanh2022MultitaskPT, Wei2022FinetunedLM, Mishra2022CrossTaskGV, Wang2022BenchmarkingGV, ouyang2022training, openai_chatgpt}. By fine-tuning an LLM with \emph{task definitions} and a few \emph{demonstration examples} on upstream training tasks, the model acquires the power to perform new tasks with unseen definitions and example. This is known as \emph{instruction learning}.

However, a natural question is: to what extent does the zero-shot generalization ability derive from the model's understanding of task definitions? Recent work in prompt-based learning suggests models might not interpret even short prompts as people expect~\citep{webson-pavlick-2022-prompt,shin2020autoprompt, deng2022rlprompt, Prasad2022GrIPS}. Task definitions are special prompts that are usually long and encode rich information. We imagine models' understanding of definitions also departs from human expectation. To investigate this question, we conduct a systematic analysis using both  human annotation and computational approaches. Our study is based on the English portion of the large-scale S{\scriptsize UPER}-N{\scriptsize ATURAL}I{\scriptsize NSTRUCTION} (NIv2) dataset~\citep{Wang2022BenchmarkingGV}, which comprises 757 training tasks and 119 unseen test tasks. 

First, we explore which type of information in task definitions is necessary for maintaining model performance. We define eight categories of content and provide a fine-grained annotation for all the sentences in task definitions. Then, we retrain the model with every occurrence of each category in NIv2 ablated out, and measure the model performance on the validation set with the same ablation. We observe variable contributions to model performance across content types. For example, input descriptions are in general not helpful to generalization performance, i.e., removing them causes little to no degradation of performance. However, larger models tend to leverage them more. On the other hand, the label information is of great importance. Providing natural-language Label Definitions helps specify the task-specific meaning of common verbalizers while providing the label verbalizer only helps in generalizing to a new label space. We also find that we can achieve similar or even better performance compared to full definitions by only providing the models with a label space along with very basic task metadata, e.g., category, domain, reasoning type, etc. This suggests that costly human generation of task definitions may not always be more helpful than available basic metadata about the task.

Second, motivated by \citet{feng2018pathologies}, to understand what is necessary for models to perform well, we propose \textbf{S}yntax-guided \textbf{T}ask \textbf{D}efinition \textbf{C}ompression (STDC), an automatic approach to removing content in task definitions that is not helpful for model performance. STDC queries the model for predictions on inputs and only requires black-box access. We can remove around 60\% of tokens while achieving \textasciitilde3 points of performance improvement of T5-XL on a held-out set. This implies that instead of understanding the whole definition of the task, the models are relying on particular text while ignoring the rest. Along with similar observations as the ablation study above, STDC reveals new patterns of how models understand definitions. For example, models usually do not need to see the whole label space, but might infer the rest with a partial label space.

Given our observations, we conclude that current instruction learning models rely on partial information in definitions. We imagine the lack of consistency in the creation process of task definitions might hinder the model from attending to all key information in definitions. Thus, we propose two complementary strategies to overcome this. The first strategy is to replace the full definition with a JSON-like formatted triplet of input, action, and output. A JSON-like triplet simplifies the creation of task definitions by asking authors of the definition to fill in blanks in templates instead of writing from scratch, and the common structure increases consistency between authors. The second strategy is to perform meta-tuning before instruction learning to adapt LLMs to any predefined styles of task definitions. We achieve 4.2, 4.0, and 2.1 Rouge-L improvements on BART-Large, T5-Large, and T5-XL, respectively, combining these two strategies. We summarize our key findings in Table \ref{table:summaryoffindings}. \footnote{Code will be released at \\ \url{https://github.com/fanyin3639/Rethinking-instruction-effectiveness}.}

\begin{figure*}
    \centering
    \includegraphics[scale=.48]{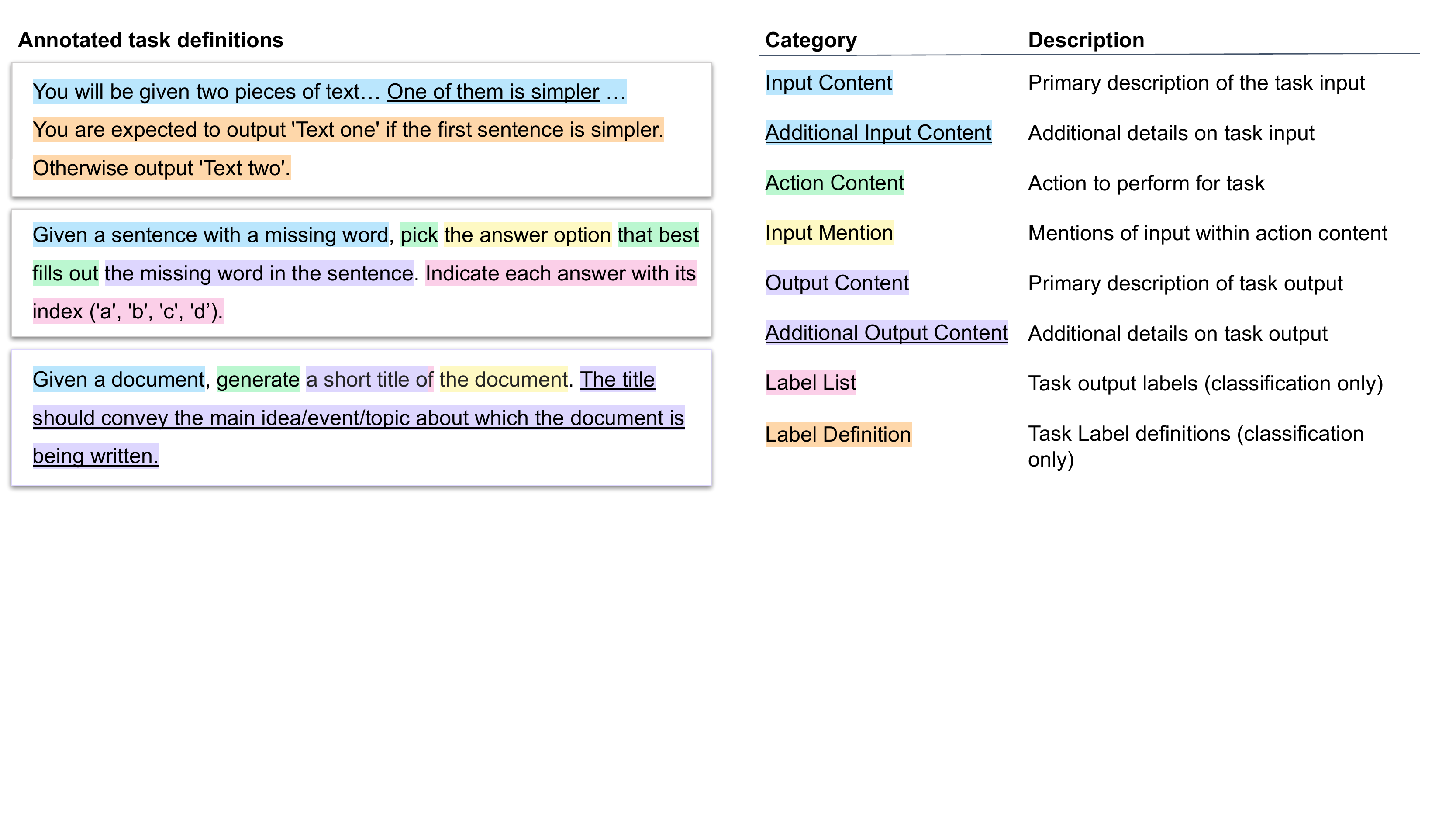}
    \vspace{-10em}
    \caption{Annotations of three examples that cover the eight categories of content in task definitions. 
    }
    \label{fig:annotation}
\end{figure*}

\section{Background}
In this section, we introduce the formulation of instruction learning, as well as the models and benchmarks used in our study. Further details are presented in Appendix \ref{appendix:details}.

\paragraph{Instruction Learning.} Instruction learning aims to train a language model so that it understands natural language task instructions and is able to generalize to a new task by solely reading new instructions. A task instruction may include several elements. In this paper, we follow~\citet{Wang2022BenchmarkingGV} and adopt instructions with 1) a \textit{task definition}: a high-level description of the input and output of the task; and 2) \textit{demonstration examples}: some input-output examples for the task. Note that other content such as \emph{things to avoid} and \emph{negative examples} may also be included but have been shown to be less effective \citep{Mishra2022CrossTaskGV}. 

A task instruction is generally pre-pended to an input and passed to the LLM. The LLM is first fine-tuned on several upstream training tasks and then asked to conduct inference on an unseen test task, given only the task instruction. 

\paragraph{Benchmark.} We adopt the English portion of NIv2~\citep{Wang2022BenchmarkingGV}, which contains 757 training tasks and 119 unseen test tasks. The test tasks fall into 12 categories, including textual entailment, data-to-text generation, etc. However, we also consider a more coarse split of test tasks into \emph{classification} and \emph{generation} tasks, based on whether the output space is fixed or not. For each task, we select 100 examples for either fine-tuning or testing and report performance of Rouge-L~\citep{lin-2004-rouge}, following~\citet{Wang2022BenchmarkingGV}. We use the task definition and two demonstration examples as the instruction. The original paper does not provide an official validation split, which we prepare by putting aside 76 training tasks. We fix the validation set for all experiments to ensure no data leakage. Note that for later experiments, results for Section \ref{section:ablation} and Section \ref{compressionsection} are reported on the validation split which we hold out ourselves while results for Section \ref{improvesection} are on the official test set.

\paragraph{Models.} We experiment with the T5-Large and T5-XL models \citep{2020t5} since the family of T5 sequence-to-sequence models has been shown by \citet{Wang2022BenchmarkingGV} to achieve superior performance after fine-tuning compared to frozen models like GPT-3~\citep{Brown2020LanguageMA} or InstructGPT~\citep{ouyang2022training} on NIv2 benchmark\footnote{At the time the paper is finished. See Section \ref{related work} and Section \ref{discussion} for updated discussions.}. We also consider BART-Large \citep{lewis2020bart} in the experiments. \textbf{All results are reported as average performance over three random seeds}. 


\section{Ablation Analysis of Annotated Task Definitions}
\label{section:ablation}


To explore what information exists in task definitions and how this impacts model performance, we manually examine all the task definitions in NIv2. We decompose and categorize definition text into eight types of content. These types cover the descriptions of input, action (the function the model should take, e.g., \textit{generate}), and output for each task in a hierarchical manner. The description can either be a primary mention of an item or provide additional, secondary details. Figure \ref{fig:annotation} shows the final categories, along with example annotations.



Three of our authors annotated all task definitions with content categories, annotating at the sentence level and in some cases sub-sentence units when required, as shown in Figure~\ref{fig:annotation}. To establish annotation feasibility, we first annotated 150 common task definitions, and measured a high inter-annotator agreement of 0.91 Fleiss Kappa~\citep{fleiss2013statistical} across categories, confirming the clarity of the defined categories. The remaining task definitions are equally split and each task is labeled by a single annotator. Appendix \ref{appendix:annotation} presents details of annotations.


\subsection{Ablation Analysis}
\label{ablationsection}
\begin{table*}[t]
	
\definecolor{LightCyan}{rgb}{0.88,1,1}
\definecolor{lightgray}{gray}{0.9}
\renewcommand{\arraystretch}{0.8}
\centering
\small
\begin{tabular}{m{1.58cm}|m{.68cm}|m{1.05cm}<{\centering}m{0.95cm}<{\centering}m{0.95cm}<{\centering}|m{1.05cm}<{\centering}m{.95cm}<{\centering}m{.95cm}<{\centering}|m{1.05cm}<{\centering}m{.95cm}<{\centering}m{.95cm}<{\centering}}
  \toprule
  &  &\multicolumn{3}{c}{\bf{BART-Large} (400M)}  & \multicolumn{3}{c}{\bf{T5-Large} (770M)} & \multicolumn{3}{c}{\bf{T5-XL} (3B)}
\cr

\cmidrule{2-5}\cmidrule{6-8}\cmidrule{9-11}
   \bf{Methods} & \%C & \bf{All}  & \bf{Cls.} & \bf{Gen.} & \bf{All}  & \bf{Cls.} & \bf{Gen.}& \bf{All}  & \bf{Cls.} & \bf{Gen.}\cr
  \toprule
  \rowcolor{lightgray}
  \multicolumn{11}{c}{Baselines} \cr
  \midrule
  \rowcolor{lightgray}
  Heuristics& - & 39.22 &53.36 &28.94 & 39.22 &53.36 &28.94& 39.22 &53.36 &28.94\cr
  \rowcolor{lightgray}
  No Def&  0\% & 38.63 & 45.77 & 33.43 &43.56 &53.52 &36.45 & 44.26&55.64 &35.99 \cr
  \rowcolor{lightgray}
  Shuffled & 100\%  & 39.73 &49.08 &32.94 &45.25 &57.17 &36.59 &48.57 &64.10 &37.26\cr
  \rowcolor{lightgray}
  Metadata& -  & 40.48 & 52.70 & 31.58 &46.79 &59.27 &37.71 &53.21 & 73.43 & 39.24\cr
  \midrule
  \rowcolor{LightCyan}
  \multicolumn{11}{c}{Full task definitions} \cr
  \midrule
  	
  \rowcolor{LightCyan}
  Full& 100\%  & 40.17 & 48.92 & 33.79 &47.55 &60.20 &38.34 &53.63 &70.82 & 41.17 \cr
  \midrule
  \multicolumn{11}{c}{Ablate Additional Information} \cr
  \midrule
  - input add& 87\%  & 40.07 & 48.84 & 33.68 &48.58&61.28 &39.26 &51.96 &67.00 &40.03 \cr
  - output add& 69\%  & 39.72 &47.62 &33.65 &48.38 &63.31 &37.51 &51.29 &66.32 &39.36 \cr
  - all add& 56\%  & 39.81 &47.90 &33.71 &48.04 &62.01 &37.89 &52.16 &66.70 &40.60 \cr
  \midrule
    \multicolumn{11}{c}{Ablate Output Information} \cr
  \midrule
    - label list& 92\%  & 36.70 & 44.23 &31.22 &44.95 &58.29 &35.26 &46.34 &60.45 &36.09 \cr
  - label desc&  89\%  & 38.04 & 47.06 & 32.10 &46.86 &57.42 &37.46 &47.25 &61.28 &37.04 \cr
  - all label&  80\%  & 36.99 &42.79 &32.78 &43.58 &55.14 &35.17 &43.85 &55.30 & 35.53 \cr
  - all output&  34\%  & 37.18 & 43.43& 32.63 &43.60 &55.24 &35.14 &43.98 &55.99 &35.23 \cr
  \midrule
  \multicolumn{11}{c}{Ablate Input Information} \cr
  \midrule
  - all input& 67\% & 39.75 &48,85 &33.14 &50.01 &64.69 &39.33 &51.61 &64.94 &41.92 \cr
  \bottomrule\hline
\end{tabular}

\caption{Comparisons of training BART-Large, T5-Large, T5-XL with full task definitions (\textcolor{cyan}{cyan}) and with other ablated alternatives. We include four baselines (\textcolor{gray}{gray}) as well as ablation experiments for certain content categories (-*). The column of \%C is the compression ratio, which refers to the fraction of remaining tokens when the content of that row is removed. We report the Rouge-L on the development task set, on all tasks (All), classification tasks (Cls.), and generation tasks (Gens.). Results show that Label information is especially important, while input information contributes marginally to the current performance.}
\label{Exp:CompareBaselines}
\end{table*}

In this section, we analyze the performance of models with ablated task definitions to understand the role of different types of information in task definitions. We also establish several baselines to better interpret the ablation results.

\paragraph{Designs of Ablations.}
We design three groups of ablation studies as follows. Note for all these ablations, we retrain the model after ablating the corresponding elements, instead of ablating at test time. Results are averaged over three random seeds.

For the first group, we remove additional information from each task definition. Additional information includes secondary information on the input and output. The ablations are as follows: \textbf{\mbox{-input add}}, which  removes all sentences marked as Additional Input Content; \textbf{-output add}, which removes all sentences marked as Additional Output Content; and \textbf{-all add}, which remove both of them.

For the second group, we ablate the output descriptions. The primary output content, i.e., the Output Content class for classification tasks includes Label List and Label Definition. Considering the importance of the label space, we design the following ablations: \textbf{-label list}, which removes all sentences marked as Label List; \textbf{-label desc}, which removes all sentences marked as Label Definition; \textbf{-all label}, which removes all label information, including both label lists and Label Definitions; and  \textbf{-all output}, which remove all sentences marked as Output Content and Additional Output Content.

For the third group, we ablate the input information. We remove all sentences marked as Input Content or Additional Input Content (\textbf{-all input}).

\paragraph{Baselines.} We consider several baselines to adequately interpret relative model performance. The \textbf{Heuristics} baseline follows similar heuristics as~\citet{Wang2022BenchmarkingGV} to serve as lower bounds of model performance. For generation tasks, this copies the input to the output. For classification tasks, it outputs a random label from the label space. The \textbf{No def} baseline removes the entire task definitions and only provides the model with the two demonstration examples. The \textbf{Shuffled} baseline provides the model with task definitions in shuffled word order. Finally, the \textbf{Metadata} baseline provides only categorical information about each task, such as its domain, reasoning type, and category, as collected by~\citet{Wang2022BenchmarkingGV}. For classification tasks, we add the label space as a metadata element. Then, we replace the original definition with a new one constructed by filling in a JSON-like template \emph{Category: \underline{1}. Reasoning type: \underline{2}. Domain: \underline{3}. Label list: {4}}, where \underline{1}, \underline{2}, \underline{3}, \underline{4} are replaced with the corresponding information for each task. Note that for generation tasks, we use ``generate free text'' to replace \underline{4}. Otherwise, \underline{4} is a comma-separated list of label verbalizers (e.g., "Yes, No").



\paragraph{Results.} Results are shown in Table \ref{Exp:CompareBaselines}. We summarize our findings from each group as follows:

\noindent {\bf Removing additional input/output information leads to little or no degradation in performance.} For all three models, we find that model performance does not change substantially after taking out the additional details of input and output, even though they contain 44\% of tokens in task definitions. However, as the model size grows, the additional information becomes slightly more influential. Removing them leads to no degradation for BART-Large and T5-Large but to a 2-point drop for T5-XL. This indicates that larger LMs can leverage the task definitions more comprehensively, another emergent ability of LLMs~\citep{wei2022emergent}.

\begin{table}[t]
\renewcommand{\arraystretch}{0.85}
\begin{center}
\small
\begin{tabular}{p{1.5cm}R{1.20cm}R{1.40cm}}
\toprule 
Label space &  \textbf{Label List} & \textbf{Label Desc.} \\
\toprule 
Seen & 0.12 & -13.21 \\
Unseen & -15.85 & -6.09 \\
\bottomrule\hline
\end{tabular}
\vspace{-6pt}
\end{center}
\caption{Performance change on classification tasks when removing Label list and Label Definitions. We take the average on two groups of dev tasks based on whether the label space has been seen during training.}
\label{table:checklabel}
\end{table}

\noindent {\bf Output content is helpful, particularly label information for classification tasks.} When removing all label information (i.e., Label List and Label Definition), model performance drops to the lowest performance, similar to having no task definition. This shows the importance of incorporating the label information in task definitions. Moreover, as the model size grows, the Label Definition has a larger positive effect on performance. It is also interesting to see removing label information causes a slight performance drop on generation tasks,  while removing all output contents, including those for generation tasks brings no further degradation. 

\noindent {\bf Input descriptions are not necessary.} Removing all direct descriptions of task inputs has nearly no negative impact on performance and leads to a slight improvement for the T5-Large model.

\noindent {\bf Comparisons with baselines.} Looking at baseline performance, we find that models with shuffled definitions usually perform better than no definition at all, indicating that token presence, even in an ungrammatical and incoherent order, can be understood by the model to some extent. Overall, the BART-Large model's performance is close to simple heuristics.
We also find that the Metadata baseline achieves similar performance as full task definitions. This provides an alternative but a far more efficient path for instruction learning, as creating structured metadata is typically less demanding than writing full natural-language task definitions. 


\subsection{The Role of Label Information}
\label{labelinfodisc}

We have shown that removing label information for classification tasks causes a substantial performance drop. We now inspect the effect of the Label List and Label Definition separately. We first split the development classification tasks into two sets: \textit{seen} verbalizers and \textit{unseen} verbalizers, based on whether the combined label verbalizers for that task appear in the training tasks. In Table~\ref{table:checklabel}, we aggregate the performance drop on these two sets when removing either the Label List or the Label Definition. We find that dropping Label List affects the performance of the unseen-verbalizer tasks most, but has no influence on the seen-verbalizer tasks. This indicates that explicitly specifying label verbalization only helps models generalize to new labels. On the other hand, dropping the Label Definitions negatively affects performance in both groups, but is more crucial in seen-verbalizer tasks. We hypothesize that models might be able to leverage the Label Definitions to disentangle the semantics of the same label names across different tasks.

\section{Compressing Task Definitions}
\label{compressionsection}

Analysis in Section~\ref{section:ablation} reveals that a large portion of information in human-written task definitions is not critical in improving model performance. This analysis is informed by human annotations. Now, to gain a model-centric perspective, we implement \textbf{S}yntax-guided \textbf{T}ask \textbf{D}efinition \textbf{C}ompression (STDC), which iteratively discovers influential content from a task definition. The motivation behind using a syntax-guided and top-down algorithm is to preserve as much human readable content as possible to show the function of compressed definitions. In our preliminary experiments, we also adopt a vanilla word-by-word compression algorithm as~\citep{feng2018pathologies}. However, we find that it is either less efficient and producing compressed definitions with slightly degraded performance on the hold-out set. 


In STDC, syntactically plausible content from the definition is iteratively removed if it does not cause a decrease in model performance. We first obtain the constituency parse tree for each definition.\footnote{With https://github.com/yzhangcs/parser} Then, in a top-down manner, we traverse the parse tree and check each phrasal node iteratively. If removing the phrase node does not cause any performance decrease, we remove the subtree rooted by that node. The algorithm stops after all leaf node removals are attempted.
The framework is illustrated in Algorithm \ref{Algo} of Appendix \ref{appendix:algorithm}.

\paragraph{Experimental Setup.}
We first train the models on the training task set with full task definitions. Then, we perform STDC during inference time on the development set for each model. The algorithm finds the compressed instruction based on a set of representative examples of task $t$, $\mathcal{D}_{t}$. To avoid over-fitting to these representatives, we test the model performance on another set of examples $\hat{\mathcal{D}}_{t}$ from the same task. We use 100 examples for both $\mathcal{D}_{t}$ and $\hat{\mathcal{D}}_{t}$.
We report the averaged Rouge-L before and after the compression, the compression ratio, i.e., the fraction of tokens in definitions being kept, and the averaged coverage score, which is the fraction of examples for which compression leads to a performance increase.
\paragraph{Results.}

\begin{table}[t]
\renewcommand{\arraystretch}{0.85}
\small
\begin{tabular}{p{1.7cm} | p{1.3cm}|p{0.8cm} p{0.6cm}p{1.1cm}}
\toprule 
\bf Model & \bf Compress. & \bf Before & \bf After  & \bf Coverage \cr
\toprule 
BART-Large & 0.52 & 40.7& 41.9 & 0.89 \cr
T5-Large &0.34 & 47.7 & 49.3  &0.92 \cr
T5-XL &0.41 & 50.3 & 53.1 & 0.89 \cr
\bottomrule
\end{tabular}
\vspace{-6pt}
\caption{Compression experiments for task definitions. We show Rouge-L results on hold-out data \textbf{Before} and  \textbf{After} compression. We also report the \textbf{Compression} ratio and averaged \textbf{Coverage} rate. Results suggest that all three models only partially understand the definitions.}
\label{compression:results}
\end{table}

\begin{figure}[t]
    \centering
    \includegraphics[scale=0.44]{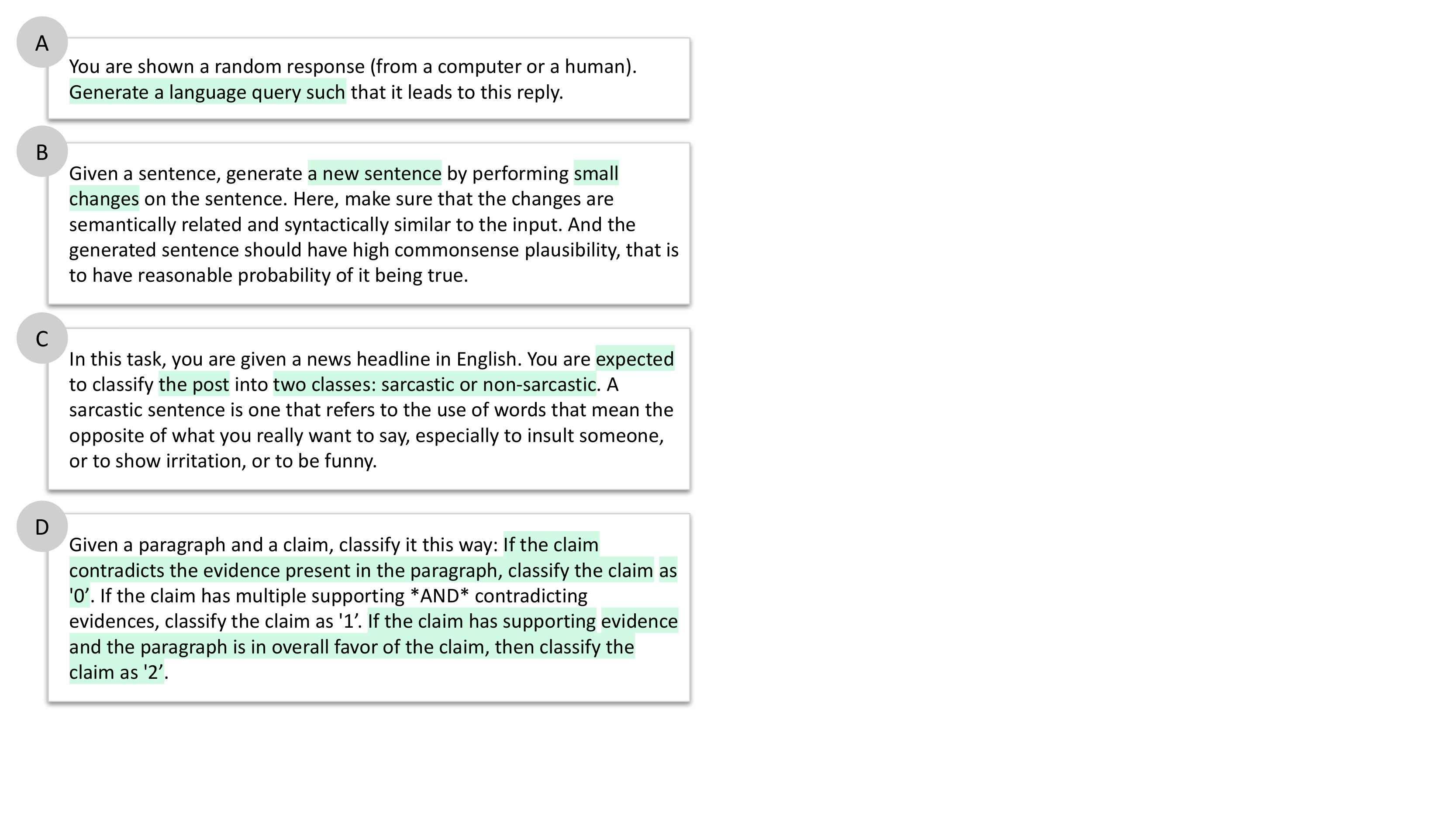}
    \vspace{-4.5em}
    \caption{Example compressions of task definitions, with retained content highlighted in green.}
    \label{fig:compressedexamples}
\end{figure}

From the results presented in Table~\ref{compression:results}, we see that for the three tested models -- BART-Large, T5-Large, and T5-XL -- we are able to remove approximately half or more of the tokens in task definitions while improving overall performance. Specifically, for T5-XL, the performance increase by 2.8 Rouge-L points while keeping only 41\% of averaged definition lengths. This echoes results in Section~\ref{ablationsection} that model performance relies on a portion of the information in task definitions. Note that the coverage averages around 90\%, indicating that the increase in performance does not come from improving outlier performance, but affects a large majority of samples. Example compressions are shown in Figure \ref{fig:compressedexamples}. We find that most compressed definitions are composed of incomplete and unnatural sentences.

\begin{figure}[t]
    \centering
    \includegraphics[scale=0.21]{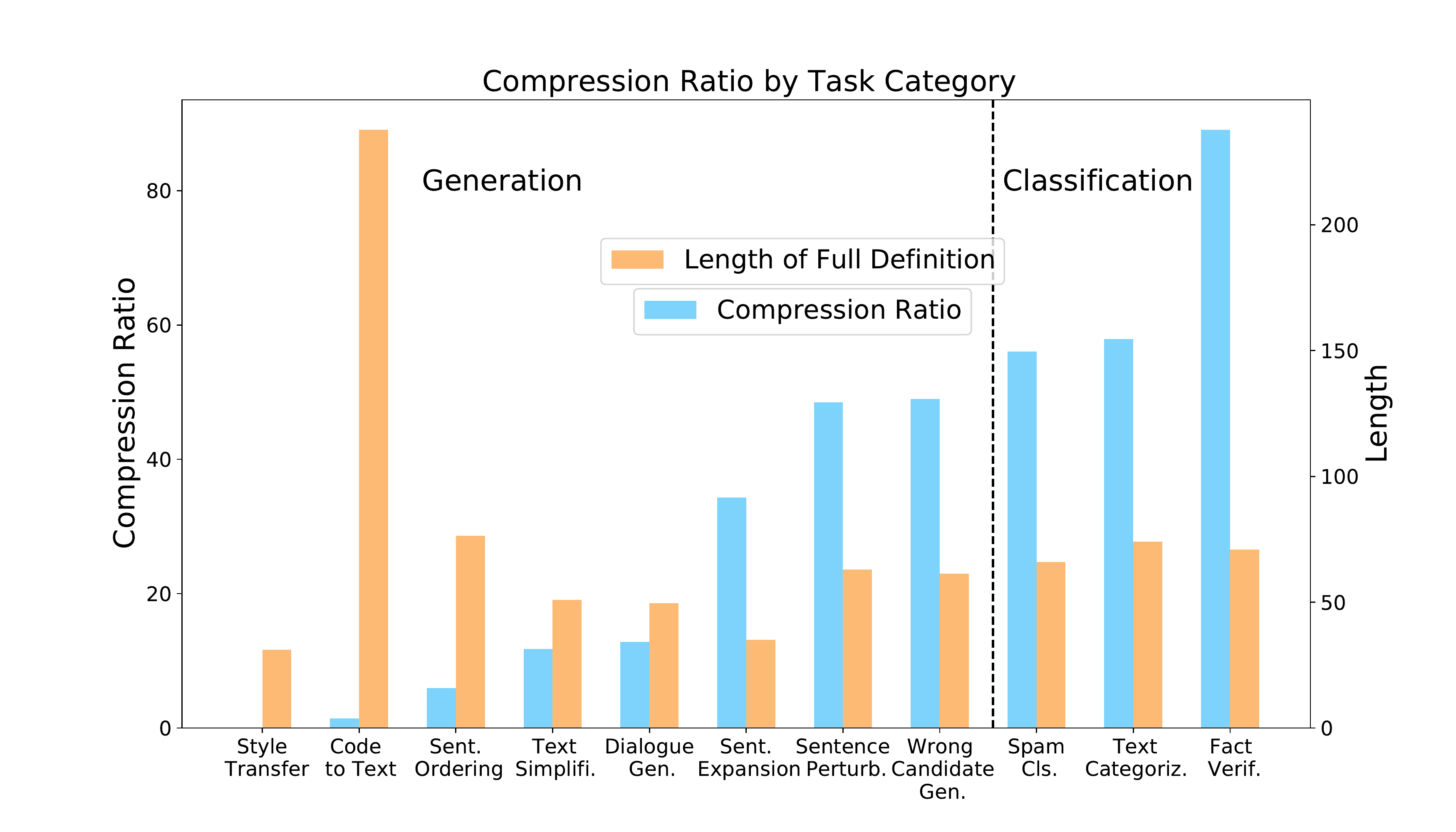}
    \caption{The compression ratio for each task category. Models tend to need less definition information for generation tasks compared to classification.}
    \label{fig:taskratio}
\end{figure}

\paragraph{Compression Ratio Distribution.} We break down the compression ratio of the STDC method by task category for the T5-XL model and show the result in Figure \ref{fig:taskratio}. Although the original definition length is roughly similar across task categories (with the exception of \textit{Code to Text}), STDC compresses significantly more content in generation tasks than in classification tasks. Two potential hypotheses are that classification tasks generally require longer task definitions, or that existing generation task definitions are not interpreted by models accurately and can be compressed extensively.

\begin{figure}[t]
    \centering
    \includegraphics[scale=0.22]{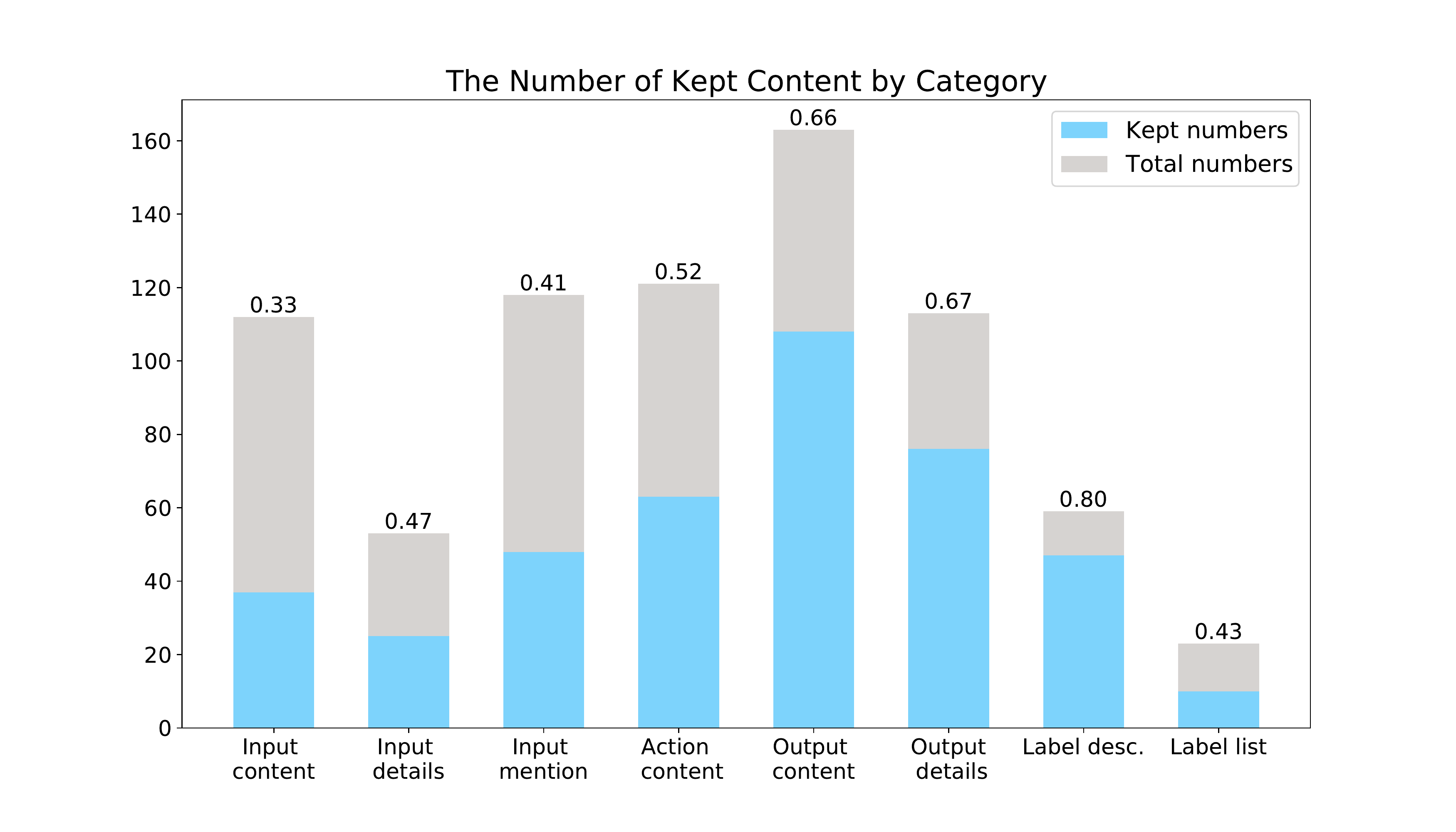}
    \caption{The number of each content category in original and compressed definitions. We put the numerical value of the fraction of kept content on top of each bar.}
    \label{fig:kept}
\end{figure}

\paragraph{Information Kept by Type} By leveraging the human annotations of information types from Section~\ref{ablationsection}, we gain insights into the information types kept after compression with STDC.
In Figure \ref{fig:kept}, we analyze the amount of content from each information type in the original task definitions compared to the amount left in the compressed instruction.


The results mirror findings in Section~\ref{ablationsection}. Specifically, 66\% of Output content and 80\% of Label Definitions are kept while only around 33\% of Input content and 47\% of Additional input details are kept, confirming that output content description is more essential than input content. The examples in Figure \ref{fig:compressedexamples} (a, b and c) illustrate this trend.

The model-centric perspective of STDC enables additional insights. Through a qualitative case study on STDC results, we find that first, only a subset of label verbalizers in the label list is required to maintain model performance, indicating that models can infer the rest of the label space based on partial labels, as shown in Figure~\ref{fig:compressedexamples}d. Second, models do not often rely on \textit{Action content}, even the root verbs, with only 52\% of the \textit{Action Content} remaining in compressed definitions. The root verbs in \textit{Action Content} are removed in examples in Figure~\ref{fig:compressedexamples}a and b, even though compressed task definition leads to better performance from the model than the full definition.

\section{Improving Model Understanding of Task Definitions}
\label{improvesection}

Previous sections indicate that not all content in task definitions contribute to strong model performance, suggesting a mismatch between the intent and model interpretation of task definitions. A possible reason for the mismatch could be due to the crowdsourcing of task definitions by many experts, creating a lack of consistency and structure in task definitions, in turn complicating the extraction of the key information by the model. To investigate the hypothesis, we propose two approaches to reduce the mismatch and improve model understanding of task definitions. First, we organize the task definition into a \emph{(input, action, output)} triplet. Second, we add a \emph{meta-tuning} stage to prepare the model before instruction learning. This phase is intended to help adapt the language models to the writing style of task definitions.

\paragraph{Structuring Task Definitions with Triplets}
We extract input/action/output information from all task definitions in NIv2 and rewrite them into triplets, leveraging both human annotation and automated processing. This serves as a starting point for using structured key information as task definitions. Future work may explore directly writing task definitions in the triplet format.


More specifically, we use a JSON-like template with the following format: \emph{Task input: \underline{1}. Task action: \underline{2}. Task output: \underline{3}}, where \underline{1}, \underline{2} and \underline{3} represent extracted portions of task definitions describing the input, action, and output, respectively. We populate the template based on the annotation we performed in Section~\ref{section:ablation}. For the input and action entries, we first extract segments marked as \emph{Input Content} and \emph{Action Content} and run a syntactic parser to extract the key phrase from the corresponding sentences.  We extract the noun phrase from \emph{Input Content} for the input entry and the verb phrase from \emph{Action Content} for the action entry. For the output entry, we use the task labels and Label Definitions for classification tasks. For generation tasks, we extract the output noun from the \emph{Action Content} sentence with rule-based methods. We manually inspected all triplets generated, manually corrected parsing mistakes, and corrected several co-reference issues we found. Some examples are presented in Appendix \ref{appendix:triplet}. Note that with this extraction process, we also fulfill the condensing of information in task definitions.

\paragraph{Meta-tuning} We also propose a meta-tuning stage specifically designed for the triplet definitions that requires the model to output entries in triplets given two demonstration examples and the entry tag. We use the same demonstration examples in the meta-tuning and instruction-learning stages of model training to avoid giving out extra data.

Specifically, during the meta-tuning stage, we provide the model with a tag \textit{[Tag]} and two demonstration examples \textit{[Example 1]} and \textit{[Example 2]}. The three options for \textit{[Tag]} are \emph{$\langle$Task input$\rangle$, $\langle$Task action$\rangle$, $\langle$Task output$\rangle$}, i.e., the keys in JSON-like triplets. Therefore, a single task triplet will split produce three training instances in the meta-tuning stage. We organize the input into a sequence of tokens: \textit{Generate segments of task definitions based on the tag and two examples. [Tag]. [Example 1]. [Example 2]}. Then, the model is trained to output the corresponding entry in task triplets for this tag with the Maximum Likelihood Estimation objective on the training task set. Finally, we initialize the parameters of instruction learning model with the meta-tuned parameters.



\begin{table}[t]
\renewcommand{\arraystretch}{0.85}
\begin{center}
\small
\begin{tabularx}{\linewidth}{p{5.40cm} p{1.30cm}}
\toprule 
Model  & \textbf{Rouge-L}\\
\toprule 
Heuristics & 38.61\\
T0 (11B) & 32.30\\
InstructGPT (175B)  & 52.10\\
\midrule
BART-Large (full def) (340M)   & 40.70{\tiny$\pm$0.4}\\
BART-Large + triplet (ours)   & 43.76{\tiny$\pm$0.3}  \\
BART-Large + triplet + meta (ours)   & \bf 44.89{\tiny$\pm$0.3} \\
\midrule
T\textit{k}-I{\scriptsize NSTRUCT}-Large (770M)   & 47.50{\tiny$\pm$0.2} \\
T\textit{k}-I{\scriptsize NSTRUCT}-Large + triplet (ours)  & 50.84{\tiny$\pm$0.1}\\
T\textit{k}-I{\scriptsize NSTRUCT}-Large + triplet + meta (ours)  & \bf 51.46{\tiny$\pm$0.2}\\
\midrule
T\textit{k}-I{\scriptsize NSTRUCT}-XL (3B) & 54.08{\tiny$\pm$0.3} \\
T\textit{k}-I{\scriptsize NSTRUCT}-XL + triplet (ours)   & 55.58{\tiny$\pm$0.2} \\
T\textit{k}-I{\scriptsize NSTRUCT}-XL + triplet + meta (ours)  & \bf 56.12{\tiny$\pm$0.2} \\
\bottomrule\hline
\end{tabularx}
\end{center}
\caption{Performances of our new strategies compared to using the standard full definitions. Standard Deviation is reported after the mean value over three random seeds.}
\label{exp:performance}
\end{table} 

\subsection{Experiments}

We compare the performance of T\textit{k}-I{\footnotesize NSTRUCT}~\cite{Wang2022BenchmarkingGV}, the state-of-the-art instruction learning model on the NIv2 benchmark, with models trained with our strategies.  T\textit{k}-I{\footnotesize NSTRUCT} is the T5 model fine-tuned on the training tasks of the benchmark. For comparisons, we also show the performance of Heuristic baselines, T0, and InstructGPT on NIv2. The results are reported on the official test set of NIv2, with 100 balanced test samples for each task. We meta-tuned the model for 10 epochs with a constant $5 \times 10^{-6}$ learning rate for BART-Large and a constant $1 \times 10^{-5}$ learning rate for T5 models, both with batch size 16. We find that the performance is not sensitive to the hyperparameters as long as we keep a small learning rate and the number of epochs under 10. Hyperparameters for instruction learning are presented in Appendix \ref{appendix:hyperp}.

\paragraph{Results} Results are summarized in Table \ref{exp:performance}. We show that both structuring task definitions with triplets and conducting the meta-tuning stage help the instruction learning performance. For the smaller models, BART-Large (340M) and T5-Large (770M), we achieve around 4 points of improvement on Rouge-L, where around 3.1 points are from structuring definitions into triplets. For the larger T5-XL (3B), we find that the structuring strategy is relatively less effective, only leading to an improvement of 1.5 points, indicating that larger models might be more effective at key information extraction from unstructured task definitions, but can still benefit from triplet formatting.

\section{Related Work}
\label{related work}
\paragraph{Instruction Learning.} Language instructions are natural ways to define tasks and easy to follow by humans. Recent works have fine-tuned pre-trained LLMs to follow instructions and generalize to new tasks with language instructions~\citep{Sanh2022MultitaskPT, Wei2022FinetunedLM, ouyang2022training, Wang2022BenchmarkingGV, chung2022scaling, openai_chatgpt, alpaca}.

\paragraph{Benchmarks of Instruction Learning.} In this work, we use the S{\scriptsize UPER}-N{\scriptsize ATURAL}I{\scriptsize NSTRUCTION} (NIv2) dataset~\citep{Wang2022BenchmarkingGV}, an enlarged task collection of \citet{Mishra2022CrossTaskGV}, which contains around 800 tasks in English with crowd-sourced instructions. Prior to this work, \citet{ye2021crossfit} test meta-learning for few-shot generalization with a collection of 160+ tasks in text-to-text format. \citet{bach2022promptsource} provide another instruction learning benchmark PromptSource with shorter and more concise task definitions. T0~\citep{Sanh2022MultitaskPT} is trained on PromptSource.

There are also recent studies that adopt automatic approaches to collect the training data of instruction learning~\citep{wang2022self, honovich2022unnatural, alpaca, peng2023instruction}. Trained models using different training data are usually evaluated on the test set of NIv2 and real user examples~\citep{wang2022self}. Our annotations on the test set of NIv2 are still useful resources for analyzing those models.

\paragraph{Prompt Engineering.} While great advance have been achieved in in-context learning~\citep{Brown2020LanguageMA} or prompt tuning~\citep{li2021prefix}, recent work has shown that we can search for better prompts by either manual engineering~\citep{schick2021few, schick2021exploiting, gao2021making, mishra2021reframing} or automatic prompt searching~\citep{shin2020autoprompt, Prasad2022GrIPS,  deng2022rlprompt}. We work with a special prompt: task definition, in the zero-shot setting. We show that better definitions can be found simply by compressing the current one. Also, we propose a new method to form definitions around structured triplets. There is also work searching for better demonstration examples~\citep{liu2022makes}, which is complementary to ours.

\paragraph{Prompt Analysis.}  Our work is most closely aligned with a line of work that analysis the role of prompts~\citep{zhao2021calibrate, webson2020do, min2022rethinking}. However, we focus on task definitions instead of short prompts or in-context examples. Also, we consider the zero-shot setting. \citet{webson2020do} find that irrelevant prompts achieve similar performance as intuitively correct prompts. We show that using metadata of a task can be comparable to using a human-written task definitions. \citet{min2022rethinking} find that label space is important for in-context learning. We further show that Label Definition can also be important, especially when needing to generalize previously seen labels in the training set to different meanings of those same labels at test time. A concurrent work with ours also analyzes the function of definitions and demonstration examples but focuses more on the label information~\citep{kung2023models}.

\section{Discussion}
\label{discussion}
The field of instruction learning has moved rapidly since this paper was first written. We summarized the newly released models and benchmarks in Section \ref{related work}. In this section, we discuss how we position the paper in the current context of instruction training, as well as how we deal with the current challenges.
\paragraph{More powerful instruction learning models} Our analysis in the previous sections is still applicable to stronger instruction learning models such as Alpaca~\citep{alpaca}. More specifically, the compression algorithm STDC can be applied to any instruction learning model to understand which part of the definitions are most useful. Moreover, since many models are still evaluated on NIv2 test set, the annotations from this paper remain relevant for continued analysis. However, we imagine that some conclusions might change. We leave this to future work and recommend people try out the resources in this paper for their own instruction learning models. Also note that no matter how the models improve, it is always important to explain how they learn to leverage instructions to do generalization, and it remains an open question.
\paragraph{Automatically created training data for instruction learning} The paradigm of prompting LLMs to generate instruction learning data has emerged as an efficient alternative to manually constructed training set. However, more efforts should be made towards improving the quality of the generated definitions under this paradigm~\citep{wang2022self}. We propose a simple method for organizing the key information in definitions. We hope later work can try combining this format with automatic instruction generations to better control the quality of data. We also notice that with the new paradigm, the boundary between content types can be vaguer than human written instructions, and there can be safety concerns regarding distilling LLMs to generate instruction tuning data~\citep{gudibande2023false}.
\paragraph{From task instructions to instructions for open-ended generation} The final goal of instruction learning is to facilitate a LLM to follow human instructions. This requires the model to advance from solving a typical NLP task like \textit{`Given a context, answer the following questions'} in a multiple-choice format, to \textit{`Tell me the procedure to book a flight ticket'}, i.e., an open-ended generation. Our analysis mainly applies to the definitions for typical NLP tasks, especially classification tasks. Later work could focus more on understanding the instructions for open-ended generations.

\section{Conclusion}
This work investigates the effectiveness of task definitions in instruction learning. Our results indicate that different types of content in definitions have widely varying impacts on model performance. Specifically, we found that label information is critical for the model performance, whereas input descriptions and additional constraints are not important. We found that current natural-language formatted definitions can be extensively compressed. We also open the door for more efficient creation of task definitions; we may simply provide the model with structured information, even the metadata, by filling in a JSON-formatted template.
\section{Limitations}
In this section, we discuss the limitations of this work. First, this study is limited to English-language tasks, due to English being the common language of the annotators. It is possible that some conclusions from this work may not extend to task definitions written in other languages; we hope that future work can extend this analysis to a multilingual context. Further, the datasets and models used may contain biases reflecting the culture of the English-speaking population, as well as biases relating to gender, race, age, and  other socio-economic factors. 

Second, in Section \ref{improvesection}, we propose a common structured format to organize the key information for a task. We rewrite the original natural language definitions into triplets after extracting key information in it and observe improved performance. However, a complementary perspective is to write such a triplet from scratch, by filling in the blanks in triplet templates and seeing whether the improvements still hold. This directly reflects whether such an organizing method works. Our approach serves as a starting point to demonstrate the effectiveness of using a structured and condensed definition.

Third, larger language models can be tested. The largest model we adopt is a T5 model with 3B parameters. As we observe variant behavior as model size grows, later work can further extend our analysis to larger models. Also, new emergent ability of LMs might be discovered with larger models, like mathematical reasoning with larger models following instructions. That is beyond the scope of this paper.

Last, some observations cannot be easily explained in this paper. For example, we saw that removing label information for classification tasks during training eventually also affects the model performance on generation tasks, which can be counter-intuitive and requires further exploration. Later work can pick a few points in the paper and provide deeper analysis on them.

\section*{Acknowledgements}
We want to thank the members of Salesforce AI Research, UCLA-NLP and UCLA PLUS-Lab for their helpful feedback and suggestions. We want to thank Prof. Kai-Wei Chang for his generous help in discussing and supporting the project. We also want to thank anonymous reviewers and chairs at ACL'23 for their invaluable comments. 





\label{ctb}
\bibliography{anthology, custom, nlp, ref}
\bibliographystyle{acl_natbib}

\clearpage
\appendix

\section{Dataset and Model Details}
\label{appendix:details}
\subsection{Validation Task set}
Since \citet{Wang2022BenchmarkingGV} do not provide an official split of the validation set, we present our own split here which is fixed across the experiments in the paper, Table \ref{table:validset} show the categories of tasks in the validation set. We find the validation tasks with the principle that there are roughly equal numbers of classification and generation tasks. The exact task names can be found in the official website \footnote{https://instructions.apps.allenai.org/}.

\begin{table}[!ht]
\renewcommand{\arraystretch}{0.85}
\begin{center}
\small
\begin{tabular}{p{4.0cm} p{1.80cm}}
\toprule 
Validation set Category &  \# Tasks \\
\toprule 
Text Categorization & 28 \\
Sentence Ordering & 3 \\
Wrong Candidate Generation & 15\\
Dialogue Generation &11 \\
Style Transfer&2 \\
Sentence Perturbation&4\\
Code to Text&4\\
Sentence Expansion&1\\
Text Simplification&4\\
Fact Verification&3\\
Spam Classification&1\\

\bottomrule\hline
\end{tabular}
\vspace{-6pt}
\end{center}
\caption{The task types in the validation set and the number of tasks in each category.}
\label{table:validset}
\end{table}

\subsection{Model Training}
T5 models and BART-Large are implemented with Huggingface's open-source library~\citep{wolf2020transformers} and the public model checkpoints \footnote{https://huggingface.co/models?sort=downloads\newline\&search=google\%2Ft5}, following the T\textit{k}-I{\footnotesize NSTRUCT} code base\footnote{https://github.com/yizhongw/Tk-Instruct}. The experiments are run on A100 GPUs with 40G memory, trained with Microsoft DeepSpeed \footnote{https://github.com/microsoft/DeepSpeed}. For all the models in Section \ref{ablationsection}, we conduct instruction learning for 2 epochs, with a constant learning rate of 5e-4, 5e-5, 1e-5, batch size 64, 32, 16  for BART-Large, T5-Large, and T5-XL, respectively. The maximum input is 1024 and the maximum output is 128. This reproduces the results in \citet{Wang2022BenchmarkingGV}.

\section{Annotation Procedure Details}
\label{appendix:annotation}
We provide details of the annotation procedure for the task definitions in NIv2 benchmark. There are in total 876 tasks in the benchmark (757 training + 119 test). Three of our authors do the annotation work on the 876 tasks. Two of them are native speakers of English. One of them is a graduate student in the United States .

\subsection{Overview of the Annotation Procedure}
To ensure the quality and objectiveness of our annotation, we adopt a three-step procedure for annotation. In the first step, the three authors look at all the task definitions and come up with a set of candidate categories. We do a trial annotation with these candidate categories on a set of randomly selected 50 tasks from the training tasks. We refine the candidate categories on these 50 task definitions until we set down with the final annotation categories. In the second step, we holdout another 150 tasks from the training tasks and everyone is asked to annotate these 150 tasks to calculate an inter-annotator agreement level. In the third step, we finish up the annotation job by equally splitting the rest tasks and assign each annotator 226 task definitions to annotate. Finally, one of the authors go through all the annotations to fix obvious errors in annotations.

\subsection{A Hierarchy of Content Types in Definitions}
We come up with the candidate categories in a hierarchical manner. We first decide the three main categories to be input, action and output descriptions. We find that these three categories cover the functionality of all the sentences in task definitions. For the input and output sentences, we further divide them into two sub-categories: Input/Output Content and Additional Input/Output Details based on whether they are primary mentions of the input/output entities or additional details or constraints.  Under the Output Content category, we create Label List and Label Definition for classification tasks, based on whether a sentence describes the semantics of the label space, or just presents a list of label verbalization. Finally, during the annotation of the first 50 task definitions, we find that sometimes the input entities will also occur in the Action Content sentence as part of the action phrase, for example, \textit{generate a summary based on the given passage}. We thus design a new class for input to refer to this special type of mentions of inputs in the Action Content sentences, named Input Mention. We do not use a `Output Mention' category because that mentions of output in Action Content is usually a primary mention of the output, which is covered by Output Content.

\begin{table}[!ht]
\renewcommand{\arraystretch}{0.85}
\begin{center}
\small
\begin{tabular}{p{4.0cm} p{1.80cm}}
\toprule 
Category &  \textbf{Agreement} \\
\toprule 
Input Content & 0.92 \\
Action Content & 0.98 \\
Output Content & 0.83 \\
Label List & 0.88 \\
Label Definition & 0.84 \\
Additional Input Details & 0.87 \\
Additional Output Details & 0.94 \\
Input Mention & 1.0 \\
\bottomrule\hline
\end{tabular}
\vspace{-6pt}
\end{center}
\caption{Performance drop on classification tasks when removing Label list and Label Definitions. We take the average on two groups of dev tasks based on whether the label space has been seen during the training time.}
\label{table:agreementlevel}
\end{table}
\subsection{Inter-Annotator Agreement Level}
We show Fleiss' kappa~\citep{fleiss2013statistical} as a statistical measurement on the agreement level of our three annotators for each category of content. Results are in Table \ref{table:agreementlevel}. The agreement level shows consistency among our annotators on all these categories, and further confirms that annotation with such  a schema is acceptable.

\subsection{Pre-process and Post-process of The Annotations}
Our annotation is in general in sentence-level. However, simply splitting a definition into sentences by the period mark is not enough for isolating the Input Content category, as the task definitions frequently use a pattern like \textit{Given a question, generate an answer...}. In this case, if we simply split at a period mark, we will get a whole sentence containing Input Content, Action content, and Output Content. For these cases, we add a rule-based pre-processing step for further splitting: we do exact match with some patterns such as \textit{Given ..., Provided with ..., and You're given ...}, and split at the next punctuation if we encounter those patterns.

After the annotations, we need to post-process the sentences marked with Action Content to extract Input Mention and Output Content if any. We do a syntactic parser on Action Content sentences and extract the root verb and its verb phrase. Then, we do another round of human annotation to mark Input Mention and Output Content within that.

\section{Compression Algorithm}
\label{appendix:algorithm}
We present the pseudo-code for the compression algorithm.
\begin{algorithm}[t]
    \small
    \caption{STDC}
    \textbf{Input:} A model $f$. a set of examples for a specific task $S$: $\mathcal{D}_{S}$. The full task definition: $X_{full} = \{w_1, w_2, ..., w_n\}$. The performance of $f$ on $\mathcal{D}_{S}$ with $x_{full}$: $f\left(\mathcal{D}_{S} | X_{full} \right)$.  Constituency tree for the task definition: $\mathcal{T}$. \\
    \textbf{Output:} Compressed definition $X_{compressed}$. 
    
    \begin{algorithmic}[1]
    \State Initialization: traverse the parse tree $\mathcal{T}$. Find the tree depth $Dep(\mathcal{T})$. The set of nodes $N_i$ at each layer i = 1,2, $\cdots$, $Dep(\mathcal{T})$.
    \State $X_{compressed}$ = $X_{full}$
    \For{layer i in 1, 2, $\cdots$, $Dep(\mathcal{T})$}
        \For{each node $n_i$ in $N_i$}
            \State  Remove $n_i$ and compute the new performance of\Statex\hspace{\algorithmicindent} \hspace{\algorithmicindent} $f$ with $X_{full} \textbackslash n_i$: $f\left(\mathcal{D}_{S} | X_{full} \textbackslash n_i \right)$
            \If {$f\left(\mathcal{D}_{S} | X_{full} \textbackslash n_i \right) \geq f\left(\mathcal{D}_{S} | X_{full} \right) $ }
            \Statex\hspace{\algorithmicindent} \hspace{\algorithmicindent} \hspace{\algorithmicindent}
            Remove $n_i$ and its subtree.
            \Statex\hspace{\algorithmicindent} \hspace{\algorithmicindent} \hspace{\algorithmicindent}$X_{compressed}$ = $X_{compressed} \textbackslash n_i$\EndIf
        \EndFor
    \EndFor
    \State  \Return $X_{compressed}$
    \end{algorithmic}
    \label{Algo}
\end{algorithm}

\section{Examples of Triplet}
\label{appendix:triplet}
We present examples of the input/action/output triplets as task definitions in Table \ref{table:examplesoftriplet}.

\section{Hyperparameters Tuning for Instruction Learning after Meta-Tuning}
\label{appendix:hyperp}
\begin{table}[h]
\renewcommand{\arraystretch}{0.85}
\begin{center}
\small
\begin{tabular}{p{2cm} p{3.80cm}}
\toprule 
Hyperparameter &  Range \\
\toprule 
Learning rate & $1 \times 10^{-5}$, $5 \times 10^{-5}$, $1 \times 10^{-4}$ \\
Batch size & 16, 32, 64, 128 \\
\# Epochs & 2, 3, 6 \\
\bottomrule\hline
\end{tabular}
\vspace{-6pt}
\end{center}
\caption{Hyper-parameter choices}
\label{table:hyperil}
\end{table}
We find that we need slightly different hyperparameters for instruction learning after meta tuning. We conduct grid search of learning rate, batch size, and the number of epochs on the validation set. The list of candidates are shown in Table \ref{table:hyperil}. The learning rate is scheduled with constant learning rate with warm up steps 300 for all models.

\begin{table*}[t]
\renewcommand{\arraystretch}{0.85}
\begin{center}
\small
\begin{tabular}{p{15.15cm}}
\toprule
\textbf{TASK1} task1292\_yelp\_review\_full\_text\_categorization \cr
\midrule
You are given a review about a place. You need to provide a rating from "1 star" to "5 stars" for this place.\cr
\midrule
\textbf{Input}: ["a review about a place"] \cr
\textbf{Action}: ["provide a rating from " 1 star " to " 5 stars " for this place"] \cr
\textbf{Output}: ["a rating from " 1 star " to " 5 stars ""] \cr
\toprule

\toprule
\textbf{TASK2} task321\_stereoset\_classification\_religion \cr
\midrule
Given two sentences regarding the stereotypes of religions, determine whether the second sentence is either a stereotype, anti-stereotype, or is unrelated with respect to the first sentence and the target. A stereotype is a widely held but oversimplified notion of a group of people. An anti-stereotype is an idea that goes against the widely held but oversimplified notion of a group of people. Sentences and targets are unrelated if there is no relevance between them.\cr
\midrule
\textbf{Input}: ["two sentences regarding the stereotypes of religions"] \cr
\textbf{Action}: ["determine whether the second sentence is either a stereotype , anti-stereotype , or is unrelated with respect to the first sentence and the target"] \cr
\textbf{Output}: ["stereotype, anti-stereotype", "A stereotype is a widely held but oversimplified notion of a group of people", "An anti-stereotype is an idea that goes against the widely held but oversimplified notion of a group of people"] \cr
\toprule
\textbf{TASK3} task628\_xlwic\_word\_with\_different\_meaning\_sentence\_generation \cr
\midrule
In this task, you are given a word, followed by a sentence. You should respond with a valid sentence which contains the word with the same meaning as in the given sentence. For example, if the given sentence refers to a 'fly' as the insect, you should not respond with a sentence which uses 'fly' as the verb. You may use the word in a different tense than is given. For example, you may use the word 'ended' in the output where the given input word is 'end'.\cr
\midrule
\textbf{Input}: ["a word, followed by a sentence"] \cr
\textbf{Action}: ["respond with a valid sentence which contains the word with the same meaning as in the given sentence"] \cr
\textbf{Output}: ["a valid sentence"] \cr
\toprule

\textbf{TASK4} task405\_narrativeqa\_question\_generation \cr
\midrule
You will be given a summary of a story. You need to create a question that can be answered from the story. You can create a question about characters, events, facts and beliefs, etc. Your question should be specific, try not to use pronouns instead of full names. As the stories are sometimes movie plots, they will contain actor names in parentheses. You should not use those names. Only use character names. Try to ask a question about all parts of the plot, not just the beginning.\cr
\midrule
\textbf{Input}: ["a summary of a story"] \cr
\textbf{Action}: ["create a question that can be answered from the story"] \cr
\textbf{Output}: ["a question"] \cr
\toprule

\textbf{TASK5} task1202\_atomic\_classification\_xneed \cr
\midrule
In this task, you are given two phrases: Head and Tail, separated with <sep>. The Head and the Tail events are short phrases possibly involving participants. The names of specific people have been replaced by generic words (e.g., PersonX, PersonY, PersonZ). PersonX is always the subject of the event. You have to determine whether it is plausible for the Head to desire the Tail or not. In this task, desire means desires of sentient entities. For example, doctors likely desire to cure a patient. Classify your answers into "Yes" and "No". The phrase may also contain a placeholder that can be an object, a person, and/or an action.\cr
\midrule
\textbf{Input}: ["two phrases : Head and Tail , separated with < sep >"] \cr
\textbf{Action}: ["determine whether it is plausible for the Head to desire the Tail or not"] \cr
\textbf{Output}: ["Yes, No"] \cr
\toprule

\textbf{TASK6} task1580\_eqasc-perturbed\_question\_generation \cr
\midrule
Given a statement, generate a question such that the answer is contained in that statement.\cr
\midrule
\textbf{Input}: ["a statement"] \cr
\textbf{Action}: ["generate a question such that the answer is contained in that statement"] \cr
\textbf{Output}: ["a question"] \cr
\toprule

\textbf{TASK7} task383\_matres\_classification \cr
\midrule
You will be given a context and a verb separated with a newline character, and you have to answer if the given verb is a negation or not. A verb is a negation if it is not going to exist, not happen, or has no effect. The output should be \"Yes\" if the verb is a negation and \"No\" otherwise.\cr
\midrule
\textbf{Input}: ["a context and a verb separated with a newline character"] \cr
\textbf{Action}: ["answer if the given verb is a negation or not"] \cr
\textbf{Output}: ["Yes, No", "" Yes " if the verb is a negation and " No " otherwise"] \cr

\bottomrule\hline
\end{tabular}
\vspace{-6pt}
\end{center}
\caption{Example of triplets.}
\label{table:examplesoftriplet}
\end{table*}

\end{document}